\documentclass[twoside]{article}

\usepackage[accepted]{aistats2026}
\usepackage[round]{natbib}

\usepackage[utf8]{inputenc} 
\usepackage[T1]{fontenc}    
\usepackage{hyperref}       
\usepackage{url}            
\usepackage{booktabs}       
\usepackage{amsfonts}       
\usepackage{nicefrac}       
\usepackage{microtype}      
\usepackage{xcolor}         

\usepackage{soul}
\usepackage[small]{caption}
\usepackage{graphicx}
\usepackage{amsmath}
\usepackage{amsthm}
\usepackage{algorithm}
\usepackage{algorithmic}
\usepackage[switch]{lineno}
\usepackage{subcaption}
\usepackage{tikz-cd}
\usetikzlibrary{arrows}
\usepackage{comment}
\usepackage{amssymb}
\usepackage{tikz}
\usepackage{tikz-qtree,tikz-qtree-compat}
\usetikzlibrary{positioning}
\usepackage{float}

\usepackage{amsmath}
\usepackage{amsthm}
\usepackage{tikz}
\usepackage{tikz-qtree,tikz-qtree-compat}

\makeatletter
\newtheorem*{rep@theorem}{\rep@title}
\newcommand{\newreptheorem}[2]{%
\newenvironment{rep#1}[1]{%
 \def\rep@title{#2 \ref{##1}}%
 \begin{rep@theorem}}%
 {\end{rep@theorem}}}
\makeatother

\newtheorem{theorem}{Theorem}
\newreptheorem{theorem}{Theorem}

\usetikzlibrary{positioning}
\newtheorem{definition}[theorem]{Definition}

\usepackage{stackengine}
\def\delequal{\mathrel{\ensurestackMath{\stackon[1pt]{=}{\scriptstyle\Delta}}}}

\begin{document}

%

%

\twocolumn[

\aistatstitle{$\epsilon$-Identifiability of Causal Quantities}

\aistatsauthor{ Ang Li \And Scott Mueller \And  Xin Shu \And Judea Pearl }

\aistatsaddress{ Florida State University \And  Toyota Research Institute \And Florida State University \And UCLA } ]


\begin{abstract}
  Identifying the effects of causes and causes of effects is vital in virtually every scientific field. Often, however, the needed probabilities may not be fully identifiable from the available data sources. This paper shows how approximate identifiability is still possible for several probabilities of causation. We term this $\epsilon\text{-identifiability}$ and demonstrate its usefulness in cases where the behavior of certain subpopulations can be restricted within sufficiently narrow bounds. In particular, we show how unidentifiable causal effects and counterfactual probabilities can be $\epsilon\text{-identified}$ when such allowances are made. Often, these allowances are easily measured and reasonably assumed. Finally, $\epsilon\text{-identifiability}$ is applied to the unit selection problem.
\end{abstract}

\section{INTRODUCTION}
Both Effects of Causes (EoC) and Causes of Effects (CoE) play an important role in many fields, such as health science, social science, and business. This is due, in part, to several features of EoC and CoE analysis. First, causal effects identified by the adjustment formula \citep{pearl1993aspects} help decision-makers avoid randomized controlled trials and use purely observational data. Second, probabilities of causation have been proven critical in policy-making and personalized decision-making \citep{mueller:pea-r513}. Third, a linear combination of probabilities of causation has been used to solve the unit selection problem defined by \citet{li:pea19-r488,li2022unit,li:pea22-r517}. Additionally, causal quantities can increase the accuracy of machine learning models by combining causal quantities with the model's label \citep{li2020training}.


The identification of causal quantities has been studied for decades. Pearl first defined causal quantities such as causal effects \citep{pearl1993aspects}, probability of necessity and sufficiency (PNS), probability of sufficiency (PS), and probability of necessity (PN) \citep{pearl1999probabilities} and their identifiability \citep{pearl2009causality} using the structural causal model (SCM) \citep{galles1998axiomatic,halpern2000axiomatizing}. Pearl also developed identification conditions of causal effects, such as back-door and front-door criteria \citep{pearl1993aspects}, along with the \emph{do}-calculus mathematically complete axiomatic identification system \citep{pearl_imaging_1994}. Pearl, Shpitser, and Bareinboim have studied more conditions for identifying causal effects \citep{shpitser:pea09-r349,bareinboim:pea12-r397}. If causal effects are not identifiable, informative bounds can be obtained. \citet{balke1997bounds} bounded causal effects from a randomized controlled trial (RCT) with imperfect compliance. \citet{li2022bounds} bounded causal effects when adjustment variables are partially observed. \citet{tian2000probabilities} proposed monotonicity as an identification condition of the binary probabilities of causation. \citet{tian2000probabilities} additionally derived tight bounds, using linear programming, when probabilities of causation are not identifiable \citep{balke1997probabilistic}. \citet{pearl:etal21-r505}, as well as \citet{dawid2017}, narrowed those bounds using additional covariate information and the corresponding causal structure. Recently, \citet{li:pea-r516} proposed theoretical work for non-binary probabilities of causation. \citet{zhang2022counterfactual} proposed numerical bounds for non-binary probabilities of causation and named it ``Partial identification'' (the difference between partial identification and the proposed $\epsilon$-Identifiability will be discussed in the conclusion section). Furthermore, \citet{shpitser:pea09-r349L} introduced another essential causal measure, known as the effect of treatment on the treated (ETT), within the SCM framework and extensively investigated its identification.

In real-world applications, decision-makers are more willing to work with identifiable cases (i.e., the causal quantities have point estimations) because the bounds under unidentifiable cases may be less informative (e.g., $0.1 \le \text{PNS} \le 0.9$). Besides, estimating the bounds often requires a combination of experimental and observational data. However, the identifiability is often hard to achieve, so we wonder if something is sitting between the identifiable and the bounds. Inspired by the idea of the confidence interval, in this paper, we proposed the definition of $\epsilon$-identifiability, in which more conditions of $\epsilon$-identifiability can be found while the estimations of the causal quantities are still near point estimations.

\section{PRELIMINARIES}
\label{related work}
\label{related work}
Here, we review the definition of ETT, PNS, PS, and PN defined by Pearl \citep{pearl1999probabilities,shpitser:pea09-r349}, as well as the definition of identifiability and the conditions for identifying ETT, PNS, PS, and PN \citep{tian2000probabilities,shpitser:pea09-r349L}. Besides, we review the tight bounds of PNS, PS, and PN when they are unidentifiable \citep{tian2000probabilities}. Readers who are familiar with the above knowledge may skip this section.

Similarly to any works mentioned above, we used the causal language of the SCM \citep{galles1998axiomatic,halpern2000axiomatizing}. The introductory counterfactual sentence ``Variable $Y$ would have the value $y$, had $X$ been $x$'' in this language is denoted by $Y_x=y$, and shorted as $y_x$. We have two types of data: experimental data, which is in the form of causal effects (denoted as $P(y_x)$), and observational data, which is in the form of a joint probability function (denoted as $P(x, y)$). Besides, in the absence of further specification, let $X$ and $Y$ be two binary variables in a causal model $M$, let $x$ and $y$ stand for the propositions $X=true$ and $Y=true$, respectively, and $x'$ and $y'$ for their complements. For notational simplicity, we limit the discussion to binary treatments and effects; extensions to multi-valued variables are discussed by Pearl (see \citep{pearl2009causality} p. 286, footnote 5).

First, the definition of identifiable for any causal quantities defined using SCM is as follows:

\begin{definition}[Identifiability]
Let $Q(M)$ be any computable quantity of a class of SCM $\mathbf{M}$ that is compatible with graph $G$. We say that $Q$ is identifiable in $\mathbf{M}$ if, for any pairs of models $M_1$ and $M_2$ from $\mathbf{M}$, $Q(M_1)=Q(M_2)$ whenever $P_{M_1}(v)=P_{M_2}(v)$, where $P(v)$ is the statistical distribution over the set $V$ of observed variables. If our observations are limited and permit only a partial set $F_M$ of features (of $P_M(v)$) to be estimated, we define $Q$ to be identifiable from $F_M$ if $Q(M_1)=Q(M_2)$ whenever $F_{M_1}=F_{M_2}$. \citep{pearl2009causality}
\end{definition}

Second, the definitions of four binary probabilities of causation defined using SCM are as follows \citep{pearl1999probabilities,shpitser:pea09-r349}:

\begin{definition}[Effect of treatment on the treated (ETT)]
\label{defett}
Let $X$ and $Y$ be two binary variables in a causal model $M$, let $x$ and $y$ stand for the propositions $X=true$ and $Y=true$, respectively, and $x'$ and $y'$ for their complements. The effect of treatment on the treated is defined as the expression 
\begin{eqnarray}
\text{ETT}&\delequal&  P(y_{x}|x') \nonumber
\end{eqnarray}
\end{definition}

\begin{definition}[Probability of necessity (PN)]
Let $X$ and $Y$ be two binary variables in a causal model $M$, let $x$ and $y$ stand for the propositions $X=true$ and $Y=true$, respectively, and $x'$ and $y'$ for their complements. The probability of necessity is defined as the expression 
\begin{eqnarray}
\text{PN} &\delequal&  P(y'_{x'}|x,y) \nonumber
\end{eqnarray}
\end{definition}

\begin{definition}[Probability of sufficiency (PS)]
Let $X$ and $Y$ be two binary variables in a causal model $M$, let $x$ and $y$ stand for the propositions $X=true$ and $Y=true$, respectively, and $x'$ and $y'$ for their complements. The probability of sufficiency is defined as the expression
\begin{eqnarray}
\text{PS} &\delequal& P(y_x|x',y') \nonumber
\end{eqnarray}
\end{definition}

\begin{definition}[Probability of necessity and sufficiency (PNS)] Let $X$ and $Y$ be two binary variables in a causal model $M$, let $x$ and $y$ stand for the propositions $X=true$ and $Y=true$, respectively, and $x'$ and $y'$ for their complements. The probability of necessity and sufficiency is defined as the expression
\begin{eqnarray}
\text{PNS}&\delequal& P(y_x,y'_{x'}) \nonumber
\end{eqnarray}
\end{definition}

Third, we review the identification conditions for causal effects \citep{pearl1993aspects,pearl1995causal}.

\begin{definition}[Back-door criterion]
Given an ordered pair of variables $(X,Y)$ in a directed acyclic graph $G$, a set of variables $Z$ satisfies the back-door criterion relative to $(X,Y)$, if no node in $Z$ is a descendant of $X$, and $Z$ blocks every path between $X$ and $Y$ that contains an arrow into $X$.
\end{definition}
If a set of variables $Z$ satisfies the back-door criterion for $X$ and $Y$, the causal effects of $X$ on $Y$ are identifiable and given by the adjustment formula:
\begin{eqnarray}
P(y_x) = \sum_z P(y|x,z)P(z).
\label{adjformula}
\end{eqnarray}

\begin{definition}[Front-door criterion]
A set of variables $Z$ is said to satisfy the front-door criterion relative to an ordered pair of variables $(X,Y)$ if 1) $Z$ intercepts all directed paths from $X$ to $Y$; 2) there is no back-door path from $X$ to $Z$; and 3) all back-door paths from $Z$ to $Y$ are blocked by $X$.
\end{definition}
If a set of variables $Z$ satisfies the front-door criterion for $X$ and $Y$, and $P(x,Z)>0$, then the causal effects of $X$ on $Y$ are identifiable and  given by the adjustment formula:
\begin{eqnarray*}
P(y_x) = \sum_z P(z|x)\sum_{x'} P(y|x',z)P(x').
\end{eqnarray*}

If causal effects are not identifiable, \citet{tian2000probabilities} provided the following bounds for the causal effects.
\begin{eqnarray}
P(x,y) \le P(y_x)\le 1 - P(x,y').
\label{inequ1}
\end{eqnarray}

Fourth, we review the identification conditions for ETT \citep{shpitser:pea09-r349L}.

\begin{theorem}
\label{thmett1}
If a set of variables $Z$ satisfies the back-door criterion for $X$ and $Y$, the ETT of $X$ on $Y$ are identifiable and given by the following formula:
\begin{eqnarray}
P(y_x|x') = \sum_z P(y|x,z)P(z|x').
\end{eqnarray}
\end{theorem}

\begin{theorem}
\label{thmett2}
The ETT of $X$ on $Y$ are identifiable if $P(Y_X)$ are identifiable and given by the following formula:
\begin{eqnarray}
P(y_x|x') = \frac{P(y_x)-P(x,y)}{P(x')}.
\end{eqnarray}
\end{theorem}

Finally, we review the identification conditions for PNS, PS, and PN \citep{tian2000probabilities}.

\begin{definition} (Monotonicity)
A Variable $Y$ is said to be monotonic relative to variable $X$ in a causal model $M$ iff
\begin{eqnarray*}
y'_x\land y_{x'}=\text{false}.
\end{eqnarray*}
\end{definition}

\begin{theorem}
If $Y$ is monotonic relative to $X$, then PNS, PN, and PS are all identifiable, and
\begin{eqnarray*}
PNS &=& P(y_x) - P(y_{x'}),\\
PN &=& \frac{P(y) - P(y_{x'})}{P(x,y)},\\
PS &=& \frac{P(y_x) - P(y)}{P(x', y')}.
\end{eqnarray*}
\end{theorem}

If PNS, PN, and PS are not identifiable, informative bounds are given by \citet{tian2000probabilities}.

\begin{eqnarray}
\max \left \{
\begin{array}{cc}
0, \\
P(y_x) - P(y_{x'}), \\
P(y) - P(y_{x'}), \\
P(y_x) - P(y)
\end{array}
\right \} \le
\text{PNS}\label{pnslb}\\
\min \left \{
\begin{array}{cc}
 P(y_x), \\
 P(y'_{x'}), \\
P(x,y) + P(x',y'), \\
P(y_x) - P(y_{x'}) +\\
P(x, y') + P(x', y)
\end{array} 
\right \}\ge\text{PNS}
\label{pnsub}
\end{eqnarray}

\begin{eqnarray}
\max \left \{
\begin{array}{cc}
0, \\
\frac{P(y)-P(y_{x'})}{P(x,y)}
\end{array} 
\right \} \le
\text{PN}\label{pnlb}\\
\min \left \{
\begin{array}{cc}
1, \\
\frac{P(y'_{x'})-P(x',y')}{P(x,y)}
\end{array}
\right \} \ge \text{PN}
\label{pnub}
\end{eqnarray}

\begin{eqnarray}
\max \left \{
\begin{array}{cc}
0, \\
\frac{P(y')-P(y'_{x})}{P(x',y')}
\end{array} 
\right \} \le
\text{PS}\label{pslb}\\
\min \left \{
\begin{array}{cc}
1, \\
\frac{P(y_{x})-P(x,y)}{P(x',y')}
\end{array}
\right \} \ge \text{PS}
\label{psub}
\end{eqnarray}

The identification conditions mentioned above (i.e., back-door and front-door criteria and monotonicity) are robust. However, it may still be hard to achieve in real-world applications. In this work, we extend the definition of identifiability, in which a sufficiently small interval is allowed. By the new definition, the estimates of causal quantities are still near point estimations, and more conditions for identifiability could be discovered. Again, if nothing is specified, the discussion in this paper will be restricted to binary treatment and effect (i.e., $X$ and $Y$ are binary).

\section{MAIN RESULTS}
First, we extend the definition of identifiability, which we call $\epsilon$-identifiability.

\begin{definition}[$\epsilon$-Identifiability]
Let $Q(M)$ be any computable quantity of a class of SCM $\mathbf{M}$ that is compatible with graph $G$. We say that $Q$ is $\epsilon$-identifiable in $\mathbf{M}$ (and $\epsilon$-identified to $q$) if for every $P(v)$, there exists $q$ s.t. for any model $M$ from $\mathbf{M}$, $Q(M)\in [q-\epsilon,q+\epsilon]$ whenever the statistical data $P_M(v)=P(v)$, where $P(v)$ is the statistical distribution over the set $V$ of observed variables. If our observations are limited and permit only a partial set $F_M$ of features (of $P_M(v)$) to be estimated, we define $Q$ to be $\epsilon$-identifiable from $F_M$ if $Q(m)\in [q-\epsilon,q+\epsilon]$ with statistical data $F_{M}$.
\end{definition}

With the above definition, the causal quantity is at a maximum distance of $\epsilon$ from its true value. We will use the infix operator symbol $\approx_\epsilon$ to represent its left-hand side being within $\epsilon$ of its right-hand side:
\begin{equation}
    r \approx_\epsilon q \iff r \in [q - \epsilon, q + \epsilon].
\end{equation}

The following sections explicate conditions for $\epsilon$-identifiability of causal effects, ETT, PNS, PS, and PN. The proofs of all the following theorems are provided in the appendix.

\subsection{$\epsilon$-Identifiability of Causal Effects}
The causal effect $P(Y_X)$ can be $\epsilon\text{-identified}$ with information about the observational joint distribution $P(X, Y)$. This can be seen by rewriting Equation \eqref{inequ1} as:
\begin{equation}
    P(x,y) \leqslant P(y_x) \leqslant P(x,y) + P(x').\label{eq:inequ2}
\end{equation}
Here, $P(y_x)$ is $\epsilon\text{-identified}$ to $P(x, y) + \epsilon$ when $P(x') \leqslant 2\epsilon$. This $\epsilon\text{-identification}$ indicates a lower bound of $P(x,y)$ and an upper bound of $P(x,y) + 2\epsilon$. Since $P(x') \leqslant 2\epsilon$, these bounds are equivalent to \eqref{eq:inequ2}. Notably, only $P(x,y)$ and an upper bound on $P(x')$ are necessary to $\epsilon\text{-identify}$ $P(y_x)$. This is generalized in Theorem \ref{thmce1}, without any assumptions of the causal structure.
\begin{theorem}
\label{thmce1}
The causal effect $P(Y_X)$ is $\epsilon$-identified as follows:
\begin{align}
    P(y_x) &\approx_\epsilon P(x,y) + \epsilon&&\text{if }P(x') \leqslant 2\epsilon,\label{thm10eq1}\\
    P(y'_x) &\approx_\epsilon P(x,y') + \epsilon&&\text{if }P(x') \leqslant 2\epsilon,\label{thm10eq2}\\
    P(y_{x'}) &\approx_\epsilon P(x',y) + \epsilon&&\text{if }P(x) \leqslant 2\epsilon,\label{thm10eq3}\\
    P(y'_{x'}) &\approx_\epsilon P(x',y') + \epsilon&&\text{if }P(x) \leqslant 2\epsilon.\label{thm10eq4}
\end{align}
\end{theorem}
When the complete distribution $P(X,Y)$ is known, Theorem \ref{thmce1} provides no extra precision over Equation \eqref{eq:inequ2}. Its power comes from when only part of the distribution is known and only an upper bound on $P(X)$ is available or able to be assumed. 

Knowledge of a causal structure can aid in $\epsilon\text{-identification}$, such as the structures shown in Figures \ref{fig:single-confounder} and \ref{fig:multiple-confounders}. (We will use Figure \ref{fig:single-confounder}, as the two figures are actually equivalent; see Definition 6 in \cite{li2022bounds} for the detailed proof of their equivalence.) If the full joint distribution $P(X, Y, U)$ were available, the causal effect $P(Y_X)$ could be computed directly via the backdoor adjustment formula. In the absence of the full joint distribution, Theorem \ref{thmce2} allows for $\epsilon\text{-identification}$ of $P(y_x)$ using only $P(x)$, the conditional probability $P(y|x)$, and an upper bound on $P(u)$.

\begin{theorem}
\label{thmce2}
Given the causal graph in Figure \ref{fig:single-confounder} and $P(u)\leqslant P(x)-c$ for some constant $c$, where $0 < c \leqslant P(x)$,
\begin{align}
    &P(y_x)
    \approx_\epsilon P(y|x)+\frac{P(x)-c}{2cP(x)+P(x)+c}\cdot\epsilon,\nonumber \\
    &\text{ if }P(u)\leqslant \frac{2cP(x)}{2cP(x)+P(x)+c}\cdot\epsilon.\label{eq:thmce2}
\end{align}
Specifically, if $P(x)\geqslant 0.5$ and $P(u)\leqslant\min\{0.1,\frac{4}{13}\epsilon\}$, then the causal effect $P(y_x)$ is $\epsilon$-identified to $P(y|x)+\frac{\epsilon}{13}$.
\end{theorem}

Note that $x \in \{x, x'\}$, $y \in \{y, y'\}$ (i.e., $x$ may be replaced by $x'$, and similarly for $y$), and $u \in \mathcal{U}$, where $\mathcal{U}$ denotes the (potentially multi-valued) domain of $U$, in Theorem \ref{thmce2} (e.g., you may substitute $x$ with $x'$ or $y$ with $y'$, and vice versa). The constant $c$ should be maximized subject to the constraints $c \leqslant P(x) - P(u)$ and the condition specified in Equation \eqref{eq:thmce2} for a given $\epsilon$. A larger $c$ implies that $P(y_x)$ is more closely $\epsilon$-identified with $P(y|x)$, but this must be balanced against minimizing $\epsilon$. Importantly, when $U$ takes on more values (i.e., when there are more potential confounders), it becomes more likely that one can find a value $u \in \mathcal{U}$ satisfying the conditions of the theorem.

As an example, if $P(x) \geqslant 0.5$ and $P(u) \leqslant 0.1$, then the causal effect $P(y_x)$ is $\epsilon$-identified to $P(y|x)+\frac{\epsilon}{13}$ if $P(u) \leqslant \frac{4}{13}\epsilon$.

Essentially, $P(y_x)$ is $\epsilon\text{-identified}$ as $P(y|x)$ plus some fraction of $\epsilon$ when there exists a $P(u)$ that is sufficiently small. Therefore, the causal effect $P(y_x)$ is close to $P(y|x)$ if $P(U)$ is specific (i.e., there exists a $P(u)$ that is sufficiently small). In this case, Theorem \ref{thmce2} can be advantageous over the backdoor adjustment formula to compute $P(y_x)$, even when data on $X$, $Y$, and $U$ are available, because $P(Y|X,U)$, required for the adjustment formula, is impractical to estimate with any $P(u)$ close to $0$.

\begin{figure}[ht]
    \centering
    \begin{subfigure}[b]{0.49\textwidth}
        \centering
        \begin{tikzpicture}[->,>=stealth',node distance=2cm,
            thick,main node/.style={circle,fill,inner sep=1.5pt}]
            \node[main node] (1) [label=above:{$U$}]{};
            \node[main node] (3) [below left=1cm of 1,label=left:$X$]{};
            \node[main node] (4) [below right=1cm of 1,label=right:$Y$] {};
            \path[every node/.style={font=\sffamily\small}]
                (1) edge (3)
                (1) edge (4)
                (3) edge (4);
        \end{tikzpicture}
        \caption{Single confounder $U$.}
        \label{fig:single-confounder}
    \end{subfigure}
\begin{subfigure}[b]{0.49\textwidth}
    \centering
    \begin{tikzpicture}[->,>=stealth',node distance=1.8cm and 1cm,
        thick,main node/.style={circle,fill,inner sep=1.5pt}]
        \node[main node] (u1) at (0,1) [label=above:{$U_1$}]{};
        \node[main node] (u2) at (1,1) [label=above:{$U_2$}]{};
        \node at (2,1) {$\dots$};
        \node[main node] (un) at (3,1) [label=above:{$U_n$}]{};
        \node[main node] (x) at (1,0) [label=left:$X$] {};
        \node[main node] (y) at (2.7,0) [label=right:$Y$] {};

        \path (u1) edge (x) (u1) edge (y);
        \path (u2) edge (x) (u2) edge (y);
        \path (un) edge (x) (un) edge (y);
        \path (x) edge (y);
    \end{tikzpicture}
    \caption{Multiple confounders $U_1,\ldots,U_n$.}
    \label{fig:multiple-confounders}
\end{subfigure}
    \caption{Causal graphs with confounders.}
\end{figure}

\subsection{$\epsilon\text{-Identifiability}$ of ETT}
The effect of treatment on the treated (ETT) is a crucial quantity in the field of causal inference, as defined in Definition \ref{defett}. ETT has been shown to be valuable for detecting latent heterogeneity \citep{pearl:15-r406-reprint}. Shpitser and Pearl proposed the identification of ETT through backdoor criteria, as outlined in Theorem \ref{thmett1}. Furthermore, Shpitser and Pearl demonstrated the relationship between the ETT and both experimental and observational data as shown in Theorem \ref{thmett2}. Therefore, Theorem \ref{thmce2} can be extended to incorporate ETT accordingly.

\begin{theorem}
    Given the causal graph in Figure \ref{fig:single-confounder} and $P(u)\leqslant P(x)-c$ for some constant $c$, where $0 < c \leqslant P(x) < 1$,
\begin{align}
    P(y_x|x')&
    \approx_\epsilon \frac{P(y|x)-P(x,y)}{P(x')}+\nonumber\\
    &+\frac{P(x)-c}{2cP(x)+P(x)+c}\cdot\epsilon ,\nonumber\\
    &\text{if }P(u)\leqslant \frac{2cP(x)P(x')}{2cP(x)+P(x)+c}\cdot\epsilon.\label{eq:thmett1}
\end{align}
Specifically, if $P(x) = 0.5$ and $P(u)\leqslant\min\{0.1, \frac{2}{13}\epsilon\}$, then the $P(y_x|x')$ is $\epsilon$-identified to $P(y|x)+\frac{\epsilon}{13}$.
    \label{thmett3}
\end{theorem}
Note that $P(x)$ cannot equal $1$, as this would make the conditional space $x'$ in $P(y_x \mid x')$ undefined. Additionally, it should be noted that $x \in {x, x'}$, $y \in {y, y'}$, and $u \in \mathcal{U}$, where $\mathcal{U}$ denotes the (potentially multi-valued) domain of $U$. It is also worth mentioning that both Theorems \ref{thmce2} and \ref{thmett3} apply to the multi-confounder case shown in Figure \ref{fig:multiple-confounders}, by using the equivalent transformation method described in Theorem 7 of \citep{li2022bounds}, making these two theorems quite general.

\subsection{$\epsilon\text{-Identifiability}$ of PNS}
Even though Tian and Pearl derived tight bounds on PNS \citep{tian2000probabilities}, the PNS can potentially be sufficiently narrow when taking into account particular upper-bound assumptions on causal effects or observational probabilities. This can be seen by analyzing the bounds of PNS in Equations \eqref{pnslb} and \eqref{pnsub}. Picking any of the arguments of the $\max$ function of the lower bound and any of the arguments to the $\min$ function of the upper bound, we can make a condition that the range of those two values is less than $2\epsilon$. For example, let us pick the second argument of the $\max$ function, $P(y_x) - P(y_{x'})$, and the first argument of the $\min$ function, $P(y_x)$:
\begin{align}
    P(y_{x'})=P(y_x) - [P(y_x) - P(y_{x'})] &\leqslant 2\epsilon.\label{eq:pns_cond_example}
\end{align}
Equation \eqref{eq:pns_cond_example} is the assumption and the PNS is the $\epsilon\text{-identified}$ to $\epsilon$ above the lower bound or $\epsilon$ below the upper bound:
\begin{align}
    \text{PNS} &\approx_\epsilon P(y_x) - P(y_{x'}) + \epsilon,\text{ or}\label{eq:pns_eps_ident_from_lb}\\
    \text{PNS} &\approx_\epsilon P(y_x) - \epsilon.\label{eq:pns_eps_ident_from_ub}
\end{align}
Since it is assumed that $P(y_{x'}) \leqslant 2\epsilon$, Equation \eqref{eq:pns_eps_ident_from_lb} is equivalent to Equation \eqref{eq:pns_eps_ident_from_ub}. The complete set of $\epsilon\text{-identifications}$ and associated conditions are stated in Theorem \ref{thmpns}.
\begin{theorem}\label{thmpns}
The PNS is $\epsilon$-identified as follows:
\begin{align}
    \text{PNS} &\approx_\epsilon \epsilon&\text{if }&P(y_x) \leqslant 2\epsilon,\label{thm12eq1}\\
    \text{PNS} &\approx_\epsilon \epsilon&\text{if }&P(y'_{x'}) \leqslant 2\epsilon,\label{thm12eq2}\\
    \text{PNS} &\approx_\epsilon \epsilon&\text{if }&P(x,y) + P(x',y') \leqslant 2\epsilon,\\
    \text{PNS} &\approx_\epsilon \epsilon&\text{if }&P(y_x) - P(y_{x'}) +\nonumber\\&&& P(x,y') + P(x',y) \leqslant 2\epsilon,
\end{align}
\begin{align}
    \text{PNS} &\approx_\epsilon P(y_x) - \epsilon&\text{if }&P(y_{x'}) \leqslant 2\epsilon,\\
    \text{PNS} &\approx_\epsilon P(y'_{x'}) - \epsilon&\text{if }&P(y'_x) \leqslant 2\epsilon,\\
    \text{PNS} &\approx_\epsilon P(y_x) -&&\nonumber\\& P(y_{x'}) + \epsilon&\text{if }&P(x,y') + P(x',y) \leqslant 2\epsilon,\\
    \text{PNS} &\approx_\epsilon P(y_x) - &\text{if }&P(y_{x'}) - P(y_{x}) +\nonumber\\&P(y_{x'}) + \epsilon && P(x,y) + P(x',y') \leqslant 2\epsilon,\\
    \text{PNS} &\approx_\epsilon P(x,y) + &\text{if }&P(y_{x'}) - P(y_{x}) +\nonumber\\&P(x',y') - \epsilon&& P(x,y) + P(x',y') \leqslant 2\epsilon,\\
    \text{PNS} &\approx_\epsilon P(y'_{x'}) - \epsilon&\text{if }&P(y') \leqslant 2\epsilon,\label{pnsex1}\\
    \text{PNS} &\approx_\epsilon P(y_{x}) - \epsilon&\text{if }&P(y_x) + P(y_{x'}) -\nonumber \\&&& P(y) \leqslant 2\epsilon,\\
    \text{PNS} &\approx_\epsilon P(y) - &\text{if }&P(y_x) + P(y_{x'}) -\nonumber\\&P(y_{x'}) + \epsilon&& P(y) \leqslant 2\epsilon,\\
    \text{PNS} &\approx_\epsilon P(x,y) + &\text{if }&P(x',y') + P(y_{x'}) -\nonumber\\&P(x',y') - \epsilon&& P(x',y) \leqslant 2\epsilon,\\
    \text{PNS} &\approx_\epsilon P(y) - &\text{if }&P(x',y') + P(y_{x'}) -\nonumber\\&P(y_{x'}) + \epsilon&& P(x',y) \leqslant 2\epsilon,\\
    \text{PNS} &\approx_\epsilon P(y) - &\text{if }&P(x,y') + P(y_{x}) -\nonumber\\&P(y_{x'}) + \epsilon&& P(x,y) \leqslant 2\epsilon,\\
    \text{PNS} &\approx_\epsilon P(y_{x}) - \epsilon&\text{if }&P(y) \leqslant 2\epsilon,\label{examplepns}\\
    \text{PNS} &\approx_\epsilon P(y'_{x'}) - \epsilon&\text{if }&P(y'_{x'}) - P(y_{x}) +\nonumber\\&&& P(y) \leqslant 2\epsilon,\\
    \text{PNS} &\approx_\epsilon P(y_x) - &\text{if }&P(y'_{x'}) - P(y_{x}) +\nonumber\\&P(y) + \epsilon&& P(y) \leqslant 2\epsilon,\\
    \text{PNS} &\approx_\epsilon P(x,y) + &\text{if }&P(x,y) + P(y'_{x}) -\nonumber\\& P(x',y') - \epsilon&& P(x,y') \leqslant 2\epsilon,\\
    \text{PNS} &\approx_\epsilon P(y_x) - &\text{if }&P(x,y) + P(y'_{x}) -\nonumber\\&P(y) + \epsilon&& P(x,y') \leqslant 2\epsilon,\\
    \text{PNS} &\approx_\epsilon P(y_x) - &\text{if }&P(x',y) + P(y'_{x'}) -\nonumber\\&P(y) + \epsilon&& P(x',y') \leqslant 2\epsilon.\label{thm12eq3}
\end{align}
\end{theorem}
Note that in the above theorem, eight conditions consist solely of experimental probabilities or solely of observational probabilities. This potentially eliminates the need for some types of studies, at least partially, even when estimating a counterfactual quantity such as PNS. For example, if a decision-maker knows that $P(y)$ is large ($P(y)\geqslant 0.95$), they can immediately conclude $\text{PNS} \approx_{0.025} P({y'}_{x'})-0.025$ (Equation (\ref{pnsex1})), without knowing the specific value of $P(y)$. Thus, only a control group study would be sufficient to estimate PNS.

There are real-world cases that exhibit near monotonicity, though not strictly so. For example, consider a scenario where a car is offered at a large discount (e.g., $50\%$). The proportion of customers who would have purchased the car at the original price but choose not to buy it when the discount is applied is likely small—indicating approximate monotonicity. In such cases, the PNS is not identifiable under the traditional definition. However, it becomes $\epsilon$-identifiable under the following theorem.

\begin{theorem}
    The PNS is $\epsilon$-identified to $P(y_x)-P(y_{x'})+\epsilon$ if $Y$ is nearly monotonic relative to $X$, such that, $P(y'_x,y_{x'})\le 2\epsilon$. 
    \label{thmnear1}
\end{theorem}

Note that monotonicity is defined in terms of counterfactuals; however, as discussed above, in some real-world cases, this probability can be estimated from experience or through other tests, as illustrated in \citep{mueller2023monotonicity}.

\subsection{$\epsilon\text{-Identifiability}$ of PN and PS}
Tian and Pearl derived tight bounds on PN and PS, in addition to PNS. Similar to the derivation in Theorem \ref{thmpns}, we can potentially obtain sufficiently narrow bounds by assuming upper bounds on causal effects or observational probabilities. The sets of $\epsilon\text{-identifications}$ and their associated conditions are presented in Theorems \ref{thmpn} and \ref{thmps}. Moreover, similar to Theorem \ref{thmnear1}, in cases of near monotonicity, the $\epsilon$-identifications of PN and PS are provided in Theorems \ref{thmnear2} and \ref{thmnear3}.

\begin{theorem}
\label{thmpn}
The PN is $\epsilon$-identified as follows:
\begin{align}
    \text{PN} &\approx_\epsilon \epsilon&&\text{if }P(y'_{x'})-P(x',y') \nonumber\\&&&\leqslant 2 \epsilon P(x,y),\label{thm13eq1}\\
    \text{PN} &\approx_\epsilon 1 - \epsilon&&\text{if }P(y_{x'})-P(x',y) \nonumber\\&&&\leqslant 2 \epsilon P(x,y),\label{thm13eq2}\\
    \text{PN} &\approx_\epsilon \frac{P(y)-P(y_{x'})}{P(x,y)}+\epsilon&&\text{if }P(y_{x'})-P(x',y) \nonumber\\&&&\leqslant 2 \epsilon P(x,y),\\
    \text{PN} &\approx_\epsilon \frac{P({y'}_{x'})-P(x',y')}{P(x,y)} -\epsilon&&\text{if }P(x,y') \nonumber\\&&&\leqslant 2\epsilon P(x,y),\\
    \text{PN} &\approx_\epsilon \frac{P(y)-P(y_{x'})}{P(x,y)}+\epsilon&&\text{if }P(x,y') \nonumber\\&&&\leqslant 2\epsilon P(x,y).\label{thm13eq3}
\end{align}
\end{theorem}
\begin{theorem}
\label{thmps}
The PS is $\epsilon$-identified as follows:
\begin{align}
    \text{PS} &\approx_\epsilon \epsilon&&\text{if }P(y_{x})-P(x,y) \nonumber\\&&&\leqslant 2 \epsilon P(x',y'),\label{thm14eq1}\\
    \text{PS} &\approx_\epsilon 1 - \epsilon&&\text{if }P(y'_{x})-P(x,y') \nonumber\\&&&\leqslant 2 \epsilon P(x',y'),\\
    \text{PS} &\approx_\epsilon \frac{P(y')-P(y'_{x})}{P(x',y')}+\epsilon&&\text{if }P(y'_{x})-P(x,y') \nonumber\\&&&\leqslant 2 \epsilon P(x',y'),
    \end{align}
\begin{align}
    \text{PS} &\approx_\epsilon \frac{P(y_{x})-P(x,y)}{P(x',y')} - \epsilon&&\text{if }P(x',y) \nonumber\\&&&\leqslant 2\epsilon P(x',y'),\\
    \text{PS} &\approx_\epsilon \frac{P(y')-P(y'_{x})}{P(x',y')}+\epsilon&&\text{if }P(x',y) \nonumber\\&&&\leqslant 2\epsilon P(x',y').
\end{align}
\end{theorem}

\begin{theorem}
    \label{thmnear2}
    The PN is $\epsilon$-identified to $\frac{P(y)-P(y_{x'})}{P(x,y)}+\epsilon$ if $Y$ is nearly monotonic relative to $X$, such that, $P(y'_x,y_{x'})\le 2\epsilon P(x,y)$. 
\end{theorem}

\begin{theorem}
    \label{thmnear3}
    The PS is $\epsilon$-identified to $\frac{P(y')-P(y'_{x})}{P(x',y')}+\epsilon$ if $Y$ is nearly monotonic relative to $X$, such that, $P(y'_x,y_{x'})\le 2\epsilon P(x',y')$. 
\end{theorem}

\section{EXAMPLES}
Here, we illustrate how to apply $\epsilon$-Identifiability in real applications by two simulated examples.
\subsection{Causal Effects of Medicine}
\label{firstexample}
Consider a medicine manufacturer who wants to know the causal effect of a new medicine on a disease. They conducted an observational study where $1500$ patients were given access to the medicine; the results of the study are summarized in Table \ref{tb1}. In addition, the expert from the medicine manufacturer acknowledged that family history is the only confounder of taking medicine and recovery, and the family history of the disease is extremely rare; only $1\%$ of the people have the family history.

Let $X=x$ denote that a patient chose to take the medicine, and $X=x'$ denote that a patient chose not to take the medicine. Let $Y=y$ denote that a patient recovered, and $Y=y'$ denote that a patient did not recover. Let $U=u$ denote that a patient has the family history, and $U=u'$ denote that a patient has no family history.

The simulated data in Table \ref{tb1} provides the estimates
\begin{eqnarray*}
    P(x) = \frac{1260}{1500} = 0.84,\\
    P(x') = \frac{240}{1500} = 0.16,\\
    P(y|x) = \frac{780}{1260} \approx 0.62,\\
    P(y|x') = \frac{210}{240} = 0.875.\\
\end{eqnarray*}
To obtain the causal effect of the medicine (i.e., using adjustment formula \eqref{adjformula}), we have to know the observational data associated with family history, which is difficult to obtain.

Fortunately, we have the prior that $P(u)=0.01$. Since $0.01=P(u)\le P(x)-0.8$ (let $c=0.8$) and $0.01=P(u)<\frac{2c*0.025P(x)}{2cP(x)+P(x)+c}\approx 0.0113$, we can apply Theorem \ref{thmce2} to obtain that $P(y_x)$ is $0.025$-identified to $P(y|x)+\frac{P(x)-c}{2cP(x)+P(x)+c}0.025\approx 0.62$. This means the causal effect of the medicine is very close to $0.62$ (i.e., $0.025$ close), which can not be $0.025$ far from $0.62$. 

\begin{table}
\centering
\caption{Results of an observational study with $1500$ individuals who have access to the medicine, where $1260$ individuals chose to receive the medicine and $240$ individuals chose not to.}
\begin{tabular}{|c|c|c|}
\hline
&Medicine&No medicine\\
\hline
Recovered&$780$&$210$\\
\hline
Not recovered&$480$&$30$\\
\hline
\end{tabular}
\label{tb1}
\end{table}

Or even simpler, note that $P(x)=0.84>0.5$ and $P(u)=0.01<0.1$, $P(u)=0.01<\frac{4}{13}*0.035\approx 0.0108$. We obtain that $P(y_x)$ is $0.035$-identified to $P(y|x)+\frac{0.035}{13}\approx 0.62$.

Similarly, since $0.01=P(u)\le P(x')-0.15$ (let $c=0.15$) and $0.01=P(u)<\frac{2c*0.1P(x')}{2cP(x')+P(x')+c}\approx 0.0134$, we can apply Theorem \ref{thmce2} again to obtain that $P(y_{x'})$ is $0.1$-identified to $P(y|x')+\frac{P(x')-c}{2cP(x')+P(x')+c}0.1\approx 0.878$.

In contrast, without the concept of ``$\epsilon$-Identifiability,'' we would have to apply the general bounds in equation (\ref{inequ1}), because the joint distribution $P(X,Y,U)$ is not available; therefore, the adjustment formula (\ref{adjformula}) cannot be applied. We have, by equation (\ref{inequ1}), $0.52 \leq P(y_x) \leq 0.68$ and $0.14 \leq P(y_{x'}) \leq 0.98$, making it impossible to reach a decision. However, by Theorem \ref{thmce2}, we find $P(y_x) \approx_{0.025} 0.62$ and $P(y_{x'}) \approx_{0.1} 0.878$, leading to the conclusion that the medicine is ineffective, without knowing the observational data associated with the family history.

\subsection{PNS of Flu Shot}
Consider a newly invented flu shot. After a vaccination company introduced a new flu shot, the number of people infected by flu reached the lowest point in 20 years (i.e., less than $5\%$ of people infected by flu). The government concluded that the new flu shot is the key to success. However, some anti-vaccination associations believe it is because people's physical quality increases yearly. Therefore, they all want to know how many percentages of people are uninfected because of the flu shot. The PNS of the flu shot (i.e., the percentage of individuals who would not infect by the flu if they had taken the flu shot and would infect otherwise) is indeed what they want.

Let $X=x$ denote that an individual has taken the flu shot and $X=x'$ denote that an individual has not taken the flu shot. Let $Y=y$ denote an individual infected by the flu and $Y=y'$ denote an individual not infected by the flu.

If they want to apply the bounds of PNS in Equations \eqref{pnslb} and \eqref{pnsub}, they must conduct both experimental and observational studies. However, note that $P(y)<0.05$, one could apply Equation \eqref{examplepns} in Theorem \ref{thmpns}, which PNS is $0.025$-identified to $P(y_x) - 0.025$ (i.e., PNS is very close to $P(y_x)$). Thus, according to the analysis of sample size in \citep{li:pea22-r518}, an experimental study for the treated group with a sample size of only $385$ is adequate for estimating PNS. 

\section{$\epsilon$-IDENTIFIABILITY IN UNIT SELECTION PROBLEM}
One utility of the causal quantities is the unit selection problem \citep{li:pea19-r488,li:pea22-r517}, in which Li and Pearl defined an objective causal function to select a set of individuals that have the desired mode of behavior.

Let $X$ denote the binary treatment and $Y$ denote the binary effect. According to Li and Pearl, individuals were divided into four response types: Complier (i.e., $P({y}_{x},{y'}_{x'})$), always-taker (i.e., $P({y}_{x},{y}_{x'})$), never-taker (i.e., $P({y'}_{x},{y'}_{x'})$), and defier (i.e., $P({y'}_{x},{y}_{x'})$). Suppose the payoff of selecting a complier, always-taker, never-taker, and defier is $\beta, \gamma, \theta, \delta$, respectively (i.e., benefit vector). The objective function (i.e., benefit function) that optimizes the composition of the four types over the selected set of individuals $c$ is as follows:
\begin{eqnarray*}
\label{liobj}
f(c) = \beta P(y_{x},y'_{x'}|c)+\gamma P(y_{x},y_{x'}|c)+\\ + \theta P(y'_{x},y'_{x'}|c)+\delta P(y'_{x},y_{x'}|c).
\end{eqnarray*}
Li and Pearl provided two types of identifiability conditions for the benefit function. One is about the response type such that there is no defier in the population (i.e., monotonicity). Another is about the benefits vector's relations, such that $\beta+\delta=\gamma+\theta$ (i.e., gain equality). These two conditions are helpful but still too specific and challenging to satisfy in real-world applications. If the benefit function is not identifiable, it can be bounded using experimental and observational data. Here in this paper, we extend the gain equality to the $\epsilon$-identifiability as stated in the following theorem.

\begin{theorem}
Given a causal diagram $G$ and distribution compatible with $G$, let $C$ be a set of variables that does not contain any descendant of $X$ in $G$, then the benefit function $f(c)=\beta P(y_x,y'_{x'}|c)+\gamma P(y_x,y_{x'}|c)+ \theta P(y'_x,y'_{x'}|c) + \delta P(y_{x'},y'_{x}|c)$ is $\frac{|\beta - \gamma - \theta + \delta|}{2}$-identified to $(\gamma -\delta)P(y_x|c)+\delta P(y_{x'}|c)+\theta P(y'_{x'}|c) + \frac{\beta - \gamma - \theta + \delta}{2}$.
\label{thmunit}
\end{theorem}
One critical use case of the above theorem is that decision-makers usually only care about the sign (gain or lose) of the benefit function. Decision-makers can apply the above theorem before conducting any observational study to see if the sign of the benefit function can be determined, as we will illustrate in the next section.

\subsection{Example: Non-immediate Profit}
Consider the most common example in \citep{li:pea19-r488}. A sale company proposed a discount on a purchase in order to increase the total non-immediate profit. The company assessed that the profit of offering the discount to complier, always-taker, never-taker, and defier is $\$100, -\$60, \$0, -\$140$, respectively. Let $X=x$ denote that a customer applied the discount, and $X=x$ denote that a customer did not apply the discount. Let $Y=y$ denote that a customer bought the purchase and $Y=y'$ denote that a customer did not. The benefit function is then (here $c$ denote all customers)
\begin{eqnarray*}
f(c) = 100 P(y_{x},y'_{x'}|c)-60 P(y_{x},y_{x'}|c) +\\+0 P(y'_{x},y'_{x'}|c)-140 P(y'_{x},y_{x'}|c).
\end{eqnarray*}
The company conducted an experimental study where $1500$ randomly selected customers were forced to apply the discount, and $1500$ randomly selected customers were forced not to. The results are summarized in Table \ref{tb2}. The experimental data reads $P(y_x|c)=0.6$ and $P(y_{x'}|c)=0.5$.

\begin{table}
\centering
\caption{Results of an experimental study with $1500$ randomly selected customers were forced to apply the discount, and $1500$ randomly selected customers were forced not to.}
\begin{tabular}{|c|c|c|}
\hline
&Discount&No discount\\
\hline
Bought the purchase&$900$&$750$\\
\hline
No purchase&$600$&$750$\\
\hline
\end{tabular}
\label{tb2}
\end{table}
Before conducting any observational study, one can conclude that the benefit function is $10$-identified to $-12$ using Theorem \ref{thmunit}. This result indicates that the benefit function is at most $10$ away from $-12$; thus, the benefit function is negative regardless of the observational data. The decision-maker then can easily conclude that the discount should not offer to the customers.
\section{CONCLUSION}

We introduced the concept of $\epsilon$-identifiability for causal quantities and established corresponding conditions for causal effects, ETT, PNS, PN, PS, and the unit selection problem. The idea is meant to capture situations where full identifiability is too demanding, yet ``near-identifiability'' within a controlled error margin is sufficient for making sound decisions. This relaxation is particularly valuable in applied settings, where strict conditions for identification are rarely met but practitioners still need principled ways to draw inferences.

Most of the conditions we presented, apart from Theorems \ref{thmce2} and \ref{thmett3}, can be derived directly from observational or experimental distributions. They therefore remain useful even when the underlying causal graph is only partially known. This makes the results immediately relevant to practice. At the same time, the two graph-dependent theorems provide a glimpse of how additional structural assumptions may help unlock $\epsilon$-identifiability in cases where standard approaches would fail. A natural next step is to search for purely graphical conditions that complement the distribution-based ones and together give a more complete picture.

Another promising direction is the incorporation of covariates and their causal relationships. Prior studies \citep{dawid2017,li2022unit,pearl:etal21-r505} already suggest that appropriate use of covariates can shrink bounds and sometimes restore identifiability, and it is plausible that similar benefits extend to the $\epsilon$-identifiable setting. Exploring this systematically could significantly broaden the scope of the framework.

A key insight is that $\epsilon$-identifiability is closely related to partial identification. While partial identification provides bounds when exact values are inaccessible, $\epsilon$-identifiability introduces explicit conditions under which the bounds are guaranteed to be within a tolerable $\epsilon$ margin. In this way, it formalizes the idea of ``near-identifiability'' and allows practitioners to reason about causal quantities with quantified imprecision deemed acceptable for practical purposes.

For each of the causal quantities considered, we provided several $\epsilon$-identifiable conditions based on different types of data. Our theorems not only show that bounded causal estimates exist but also give explicit ranges, thereby turning intractable questions into informative inference problems. This dual emphasis on existence and explicitness is one of the strengths of the approach.

Our main contribution is to lay the foundation for this new line of research. While tightness was not the focus here, we aimed to find minimal conditions to achieve a specified $\epsilon$. Future work could explore how to define and prove tightness for multiple overlapping conditions.

We hope this framework inspires future theoretical developments and practical applications of $\epsilon$-identifiability in causal inference.

    \section*{Acknowledgements}
    This research was supported in parts by grants from the National Science
    Foundation [\#IIS-2106908 and \#CNS-2321786], and Toyota Research Institute of North America
    [\#PO-000897].

\bibliographystyle{plainnat}
\bibliography{Ang}

\clearpage
\appendix
\thispagestyle{empty}

\onecolumn
\aistatstitle{Supplementary Materials}

\section{Technical Appendices and Supplementary Material}

\subsection{Proof of Theorem \ref{thmce1}}\label{proof:thmce1}
\begin{proof}
From Equation \eqref{inequ1} we have,
\begin{eqnarray*}
P(x,y) \le P(y_x) \le 1 - P(x,y').
\end{eqnarray*}
Let $1-P(x,y')-P(x,y) \le 2\epsilon$, we obtain $P(x')\le 2\epsilon$.\\
Therefore, $P(y_x)$ is $\epsilon$-identified to $P(x,y)+\epsilon$ if $P(x')\le 2\epsilon$, Equation \eqref{thm10eq1} holds. Similarily, we can substitute $x,y$ with $x',y'$, respectively. Equations \eqref{thm10eq2} to \eqref{thm10eq4} hold.
\end{proof}
\subsection{Proof of Theorem \ref{thmce2}}\label{proof:thmce2}
\begin{proof}
Let $u\in \mathcal{U}$, and $u' = \cup_{u''\in\mathcal{U},u''\neq u} u''$.
Then, we have,
\begin{eqnarray*}
P(y_x) &=& P(y_x|u)P(u)+P(y_x|u')P(u')\\
&=&P(y|x,u)P(u)+P(y|x,u')P(u') \text{ since u satisfied the back-door}
\end{eqnarray*}
Thus,
\begin{eqnarray*}
&&P(y_x)\\
&\ge& P(y|x,u')P(u')\\
&=& P(y|x,u')(1-P(u))\\
&=& \frac{P(x,y,u')}{P(x,u')}(1-P(u))\\
&\ge& \frac{P(x,y)-P(u)}{P(x)}(1-P(u))\\
&=& P(y|x)-P(y|x)P(u)-\frac{P(u)}{P(x)}+\frac{P^2(u)}{P(x)}\\
&\ge& P(y|x)-P(u)-\frac{P(u)}{P(x)}\\
&=& P(y|x)-(1 + \frac{1}{P(x)})P(u).
\end{eqnarray*}
Also if $P(x)\ge P(u)+c$ for some constant $c>0$, we have,
\begin{eqnarray*}
&&P(y_x)\\
&\le& P(u) + P(y|x,u')(1-P(u))\\
&\le& P(u) + \frac{P(x,y,u')}{P(x,u')}(1-P(u))\\
&\le& P(u) + \frac{P(x,y)}{P(x)-P(u)}(1-P(u))\\
&\le& P(u) + \frac{P(x,y)}{P(x)-P(u)}\\
&=& P(u) + \frac{P(x,y)}{P(x)(1-\frac{P(u)}{P(x)})}\\
&=& P(u) + \frac{P(x,y)(1-\frac{P(u)}{P(x)})+P(y|x)P(u)}{P(x)(1-\frac{P(u)}{P(x)})}\\
&=& P(u) + P(y|x)+\frac{P(y|x)P(u)}{P(x)-P(u)}\\
&\le& P(y|x) + P(u) + \frac{P(u)}{P(x)-P(u)}\\
&\le& P(y|x) + P(u) + \frac{P(u)}{c}\\
&=& P(y|x) + P(u)(1 + \frac{1}{c})\\
\end{eqnarray*}
Therefore, we have, 
\begin{eqnarray*}
P(y|x)-(1 + \frac{1}{P(x)})P(u) \le P(y_x),\\
P(y|x) + (1 + \frac{1}{c})P(u) \ge P(y_x).
\end{eqnarray*}
Let
\begin{eqnarray*}
(1 + \frac{1}{c})P(u)+(1 + \frac{1}{P(x)})P(u) \le 2\epsilon.
\end{eqnarray*}
We have,
\begin{eqnarray*}
&&P(u)\\
&\le& \frac{2}{2+\frac{1}{c}+\frac{1}{P(x)}}\epsilon\\
&=&\frac{2cP(x)}{2cP(x)+P(x)+c}\epsilon.
\end{eqnarray*}
Then we know that if $P(u)\le \frac{2cP(x)}{2cP(x)+P(x)+c}\epsilon$,
\begin{eqnarray*}
P(y|x)-(1 + \frac{1}{P(x)})\frac{2cP(x)}{2cP(x)+P(x)+c}\epsilon \le &P(y_x),\\P(y|x) + (1 + \frac{1}{c})\frac{2cP(x)}{2cP(x)+P(x)+c}\epsilon\ge &P(y_x),\\
P(y|x)-\frac{2cP(x)+2c}{2cP(x)+P(x)+c}\epsilon \le &P(y_x),\\ P(y|x) + \frac{2cP(x)+2P(x)}{2cP(x)+P(x)+c}\epsilon\ge &P(y_x).
\end{eqnarray*}
Therefore, $P(y_x)$ is $\epsilon$-identified to $P(y|x)-\frac{2cP(x)+2c}{2cP(x)+P(x)+c}\epsilon + \epsilon=P(y|x)+\frac{P(x)-c}{2cP(x)+P(x)+c}\epsilon$.\\
Besides, if $P(x)\ge 0.5$ and $P(u)\le 0.1$, let $c=0.4$, we have 
\begin{eqnarray*}
P(y|x)-(1 + \frac{1}{P(x)})P(u) \le P(y_x),&\\ P(y|x) + (1 + \frac{1}{c})P(u) \ge P(y_x).&\\
P(y|x)-(1 + \frac{1}{0.5})P(u) \le P(y_x),&\\ P(y|x) + (1 + \frac{1}{0.4})P(u) \ge P(y_x).&\\
P(y|x)-3P(u) \le P(y_x) \le P(y|x) + 3.5P(u).&
\end{eqnarray*}
Let $3.5P(u)+3P(u)\le 2\epsilon$, we have $P(u)\le \frac{4}{13}\epsilon$, and 
\begin{eqnarray*}
P(y|x)-\frac{12}{13}\epsilon \le &P(y_x)& \le P(y|x) + \frac{14}{13}\epsilon.
\end{eqnarray*}
Therefore, $P(y_x)$ is $\epsilon$-identified to $P(y|x)-\frac{12}{13}\epsilon+\epsilon=P(y|x)+\frac{\epsilon}{13}$.
\end{proof}

\subsection{Proof of Theorem \ref{thmett3}}\label{proof:thmett3}
\begin{proof}
Similar to the proof of Theorem \ref{thmce2}, let $u\in \mathcal{U}$ and $u' = \cup_{u''\in\mathcal{U},u''\neq u} u''$, we have,
\begin{eqnarray*}
P(y|x)-(1 + \frac{1}{P(x)})P(u) \le P(y_x),\\
P(y|x) + (1 + \frac{1}{c})P(u)\ge P(y_x).
\end{eqnarray*}
then we obtain,
\begin{eqnarray*}
\frac{P(y|x)}{P(x')}-(1 + \frac{1}{P(x)})\frac{P(u)}{P(x')} \le \frac{P(y_x)}{P(x')},\\\frac{P(y|x)}{P(x')} + (1 + \frac{1}{c})\frac{P(u)}{P(x')}\ge \frac{P(y_x)}{P(x')}.
\end{eqnarray*}
Let
\begin{eqnarray*}
(1 + \frac{1}{c})\frac{P(u)}{P(x')}+(1 + \frac{1}{P(x)})\frac{P(u)}{P(x')} \le 2\epsilon.
\end{eqnarray*}
We have,
\begin{eqnarray*}
&&P(u)\\
&\le& \frac{2P(x')}{2+\frac{1}{c}+\frac{1}{P(x)}}\epsilon\\
&=&\frac{2cP(x)P(x')}{2cP(x)+P(x)+c}\epsilon.
\end{eqnarray*}
Then we know that if $P(u)\le \frac{2cP(x)P(x')}{2cP(x)+P(x)+c}\epsilon$,
\begin{eqnarray*}
\frac{P(y|x)}{P(x')}-(1 + \frac{1}{P(x)})\frac{2cP(x)}{2cP(x)+P(x)+c}\epsilon \le &\frac{P(y_x)}{P(x')},\\\frac{P(y|x)}{P(x')} + (1 + \frac{1}{c})\frac{2cP(x)}{2cP(x)+P(x)+c}\epsilon\ge &\frac{P(y_x)}{P(x')},\\
\frac{P(y|x)}{P(x')}-\frac{2cP(x)+2c}{2cP(x)+P(x)+c}\epsilon \le &\frac{P(y_x)}{P(x')},\\ \frac{P(y|x)}{P(x')} + \frac{2cP(x)+2P(x)}{2cP(x)+P(x)+c}\epsilon\ge &\frac{P(y_x)}{P(x')}.
\end{eqnarray*}

By Theorem \ref{thmett2}, we have,
\begin{eqnarray*}
P(y_x|x') = \frac{P(y_x)}{P(x')}-\frac{P(x,y)}{P(x')}.
\end{eqnarray*}

Therefore, $P(y_x|x')$ is $\epsilon$-identified to $\frac{P(y|x)}{P(x')}-\frac{P(x,y)}{P(x')}-\frac{2cP(x)+2c}{2cP(x)+P(x)+c}\epsilon + \epsilon=\frac{P(y|x)}{P(x')}-\frac{P(x,y)}{P(x')}+\frac{P(x)-c}{2cP(x)+P(x)+c}\epsilon$.\\
Besides, if $P(x) = 0.5$ and $P(u)\le 0.1$, let $c=0.4$, similarly to the proof of Theorem \ref{thmce2}, we have,
\begin{eqnarray*}
P(y|x)-(1 + \frac{1}{0.5})P(u) \le P(y_x),&\\ P(y|x) + (1 + \frac{1}{0.4})P(u) \ge P(y_x).&\\
2P(y|x)-2(1 + \frac{1}{0.5})P(u) \le 2P(y_x),&\\ 2P(y|x) + 2(1 + \frac{1}{0.4})P(u) \ge 2P(y_x).&\\
2P(y|x)-6P(u) \le 2P(y_x) \le 2P(y|x) + 7P(u).&
\end{eqnarray*}

By Theorem \ref{thmett2} and $P(x)=P(x')=0.5$, we have,
\begin{eqnarray*}
P(y_x|x') &=& \frac{P(y_x)}{P(x')}-\frac{P(x,y)}{P(x')}\\
&=&\frac{P(y_x)}{0.5}-\frac{P(x,y)}{P(x)}\\
&=&2P(y_x)-P(y|x).
\end{eqnarray*}

Let $7P(u)+6P(u)\le 2\epsilon$, we have $P(u)\le \frac{2}{13}\epsilon$, and 
\begin{eqnarray*}
P(y|x)-\frac{12}{13}\epsilon \le &P(y_x|x')& \le P(y|x) + \frac{14}{13}\epsilon.
\end{eqnarray*}
Therefore, $P(y_x|x')$ is $\epsilon$-identified to $P(y|x)-\frac{12}{13}\epsilon+\epsilon=P(y|x)+\frac{\epsilon}{13}$.
\end{proof}

\subsection{Proof of Theorem \ref{thmpns}}\label{proof:thmpns}
\begin{proof}
From the bounds of PNS in Equations \eqref{pnslb} and \eqref{pnsub} is as follows:
\begin{eqnarray*}
\max \left \{
\begin{array}{cc}
0, \\
P(y_x) - P(y_{x'}), \\
P(y) - P(y_{x'}), \\
P(y_x) - P(y)
\end{array}
\right \} \le
\text{PNS}\\ \min \left \{
\begin{array}{cc}
 P(y_x), \\
 P(y'_{x'}), \\
P(x,y) + P(x',y'), \\
P(y_x) - P(y_{x'}) +\\
+ P(x, y') + P(x', y)
\end{array} 
\right \}\ge \text{PNS}.
\end{eqnarray*}
Let $P(y_x)-0\le 2\epsilon$, we obtain that PNS is $\epsilon$-identified to $\epsilon$ if $P(y_x)\le 2\epsilon$, Equation \eqref{thm12eq1} holds.\\
Similarly, the rest of $20$ equations can be obtained by letting 
\begin{eqnarray*}
P({y'}_{x'})-0&\le& 2\epsilon,\\
P(x,y) + P(x',y')-0&\le& 2\epsilon,\\
P(y_x) - P(y_{x'}) + P(x, y') + P(x', y)-0&\le& 2\epsilon,\\
P(y_x)-(P(y_x) - P(y_{x'}))&\le& 2\epsilon,\\
P(y'_{x'})-(P(y_x) - P(y_{x'}))&\le& 2\epsilon,\\
P(x,y) + P(x',y')-(P(y_x) - P(y_{x'}))&\le& 2\epsilon,\\
P(y_x) - P(y_{x'}) + P(x, y') + P(x', y)-&&\\(P(y_x) - P(y_{x'}))&\le& 2\epsilon,\\
P(y_x)-(P(y) - P(y_{x'}))&\le& 2\epsilon,\\
P(y'_{x'})-(P(y) - P(y_{x'}))&\le& 2\epsilon,\\
P(x,y) + P(x',y')-(P(y) - P(y_{x'}))&\le& 2\epsilon,\\
P(y_x) - P(y_{x'}) + P(x, y') + P(x', y)-&&\\(P(y) - P(y_{x'}))&\le& 2\epsilon,\\
P(y_x)-(P(y_x) - P(y))&\le& 2\epsilon,\\
P(y'_{x'})-(P(y_x) - P(y))&\le& 2\epsilon,\\
P(x,y) + P(x',y')-(P(y_x) - P(y))&\le& 2\epsilon,\\
P(y_x) - P(y_{x'}) + P(x, y') + P(x', y)-&&\\(P(y_x) - P(y))&\le& 2\epsilon.
\end{eqnarray*}
\end{proof}

\subsection{Proof of Thereom \ref{thmnear1}}
\label{proof:thmnear1}
\begin{proof}
First, we have,
\begin{eqnarray*}
&&P(y_x,y'_{x'}) - P(y_{x'},y'_x)\\
&=&P(y_x,y'_{x'},x) + P(y_x,y'_{x'},x') -P(y_{x'},y'_{x},x) - P(y_{x'},y'_{x},x') \\
&=&P(y,y'_{x'},x) + P(y_x,y',x') -P(y_{x'},y',x) - P(y,y'_{x},x') \\
&=&P(y,y'_{x'},x) - P(y_{x'},y',x) +P(y_x,y',x') - P(y,y'_{x},x') \\
&=&P(x,y) - P(y,y_{x'},x) - P(y_{x'},y',x) +P(y_x,y',x') + P(y,y_{x},x') - P(x',y)\\
&=&P(x,y) - P(y_{x'},x) + P(y_x,x') - P(x',y)\\
&=&P(x,y) - P(y_{x'}) + P(y_{x'},x') +P(y_x) - P(y_x,x) - P(x',y)\\
&=&P(x,y) - P(y_{x'}) + P(y,x') +P(y_x) - P(y,x) - P(x',y)\\
&=&P(y_x) - P(y_{x'})
\end{eqnarray*}
Then, we have,
\begin{eqnarray*}
&&PNS\\
&=&P(y_x,y'_{x'}) \nonumber \\
&=&P(y_x)-P(y_{x'})+P(y'_x,y_{x'})
\end{eqnarray*}
Plus if $P(y'_x,y_{x'})\le 2\epsilon$, we have,
\begin{eqnarray*}
P(y_x)-P(y_{x'})\le PNS \le P(y_x)-P(y_{x'})+2\epsilon
\end{eqnarray*}
and we obtain that,
\begin{eqnarray*}
PNS\approx_\epsilon P(y_x)-P(y_{x'})+\epsilon.
\end{eqnarray*}
\end{proof}

\subsection{Proof of Theorem \ref{thmpn}}\label{proof:thmpn}
\begin{proof}
From the bounds of PN in Equations \eqref{pnlb} and \eqref{pnub} is as follows:
\begin{eqnarray*}
\max \left \{
\begin{array}{cc}
0, \\
\frac{P(y)-P(y_{x'})}{P(x,y)}
\end{array} 
\right \} \le
\text{PN} \le
\min \left \{
\begin{array}{cc}
1, \\
\frac{P(y'_{x'})-P(x',y')}{P(x,y)}
\end{array}
\right \}
\end{eqnarray*}
Let $\frac{P(y'_{x'})-P(x',y')}{P(x,y)}-0\le 2\epsilon$, we obtain that PN is $\epsilon$-identified to $\epsilon$ if $P(y'_{x'})-P(x',y')\le 2P(x,y)\epsilon$, Equation \eqref{thm13eq1} holds.\\
Similarly, the rest of $4$ equations can be obtained by letting 
\begin{eqnarray*}
1-\frac{P(y)-P(y_{x'})}{P(x,y)}&\le& 2\epsilon,\\
\frac{P(y'_{x'})-P(x',y')}{P(x,y)} - \frac{P(y)-P(y_{x'})}{P(x,y)}&\le& 2\epsilon.
\end{eqnarray*}
\end{proof}

\subsection{Proof of Theorem \ref{thmps}}\label{proof:thmps}
\begin{proof}
From the bounds of PS in Equations \eqref{pslb} and \eqref{psub} is as follows:
\begin{eqnarray*}
\max \left \{
\begin{array}{cc}
0, \\
\frac{P(y')-P(y'_{x})}{P(x',y')}
\end{array} 
\right \} \le
\text{PS} \le
\min \left \{
\begin{array}{cc}
1, \\
\frac{P(y_{x})-P(x,y)}{P(x',y')}
\end{array}
\right \}
\end{eqnarray*}
Let $\frac{P(y_{x})-P(x,y)}{P(x',y')}-0\le 2\epsilon$, we obtain that PS is $\epsilon$-identified to $\epsilon$ if $P(y_{x})-P(x,y)\le 2P(x',y')\epsilon$, Equation \eqref{thm14eq1}.\\
Similarly, the rest of $4$ conditions can be obtained by letting 
\begin{eqnarray*}
1-\frac{P(y')-P({y'}_{x})}{P(x',y')}&\le& 2\epsilon,\\
\frac{P(y_{x})-P(x,y)}{P(x',y')} - \frac{P(y')-P({y'}_{x})}{P(x',y')}&\le& 2\epsilon.
\end{eqnarray*}
\end{proof}

\subsection{Proof of Thereom \ref{thmnear2}}
\label{proof:thmnear2}
\begin{proof}
\begin{eqnarray*}
&&PN\\
&=&P(y'_{x'}|x,y)\\
&=&\frac{P(y'_{x'},x,y)}{P(x,y)}\\
&=&\frac{P(y'_{x'})-P(x',y')-P(y'_{x'},x,y')}{P(x,y)}\\
&=&\frac{P(y'_{x'})-P(x',y')-P(x,y')+P(y_{x'},x,y')}{P(x,y)}\\
&=&\frac{P(y'_{x'})-P(y')+P(y_{x'},x,y'_{x})}{P(x,y)}\\
&=&\frac{P(y)-P(y_{x'})+P(y_{x'},y'_{x},x)}{P(x,y)}
\end{eqnarray*}
Plus if $P(y'_x,y_{x'})\le 2\epsilon P(x,y)$, we have,
\begin{eqnarray*}
    \frac{P(y'_x,y_{x'},x)}{P(x,y)}\le \frac{P(y'_x,y_{x'})}{P(x,y)} \le 2\epsilon
\end{eqnarray*}
and then,
\begin{eqnarray*}
\frac{P(y)-P(y_{x'})}{P(x,y)}\le PN \le \frac{P(y)-P(y_{x'})}{P(x,y)}+2\epsilon
\end{eqnarray*}
and we obtain that,
\begin{eqnarray*}
PN\approx_\epsilon \frac{P(y)-P(y_{x'})}{P(x,y)}+\epsilon.
\end{eqnarray*}
\end{proof}

\subsection{Proof of Thereom \ref{thmnear3}}
\label{proof:thmnear3}
\begin{proof}
    The proof follows exactly the same steps as in the proof of Theorem \ref{thmnear2} in the previous section, except that $x$ is replaced by $x'$, $x'$ by $x$, $y$ by $y'$, and $y'$ by $y$.
\end{proof}
\subsection{Proof of Theorem \ref{thmunit}}
\begin{proof}
\begin{eqnarray}
&&f(c)\nonumber\\
&=&\beta P(y_x,y'_{x'}|c)+\gamma P(y_{x},y_{x'}|c)+\nonumber\\
&&\theta P(y'_x,y'_{x'}|c)+\delta P(y'_{x},y_{x'}|c)\nonumber \\
&=&\beta P(y_x,y'_{x'}|c)+\gamma [P(y_x|c)-P(y_{x},y'_{x'}|c)]+\nonumber\\
&&\theta [P(y'_{x'})-P(y_x,y'_{x'}|c)]+\delta P(y'_{x},y_{x'}|c)\nonumber \\
&=&\gamma P(y_x|c)+\theta P(y'_{x'}|c)+(\beta -\gamma -\theta) P(y_x,y'_{x'}|c)+\nonumber\\
&&\delta P(y'_{x},y_{x'}|c).
\label{eqb111}
\end{eqnarray}
Note that, we have,
\begin{eqnarray}
P(y'_x,y_{x'}|c)=P(y_x,y'_{x'}|c)-P(y_x|c)+P(y_{x'}|c).
\label{eqb222}
\end{eqnarray}
Substituting Equation \eqref{eqb222} into Equation \eqref{eqb111}, we have,
\begin{eqnarray}
&&f(c)\nonumber\\
&=&\gamma P(y_x|c)+\theta P(y'_{x'}|c)+(\beta -\gamma -\theta) P(y_x,y'_{x'}|c)+\nonumber\\
&&\delta P(y'_{x},y_{x'}|c)\nonumber\\
&=&\gamma P(y_x|c)+\theta P(y'_{x'}|c)+(\beta -\gamma -\theta) P(y_x,y'_{x'}|c)+\nonumber\\
&&\delta [P(y_x,y'_{x'}|c)-P(y_x|c)+P(y_{x'}|c)]\nonumber\\
&=&(\gamma -\delta)P(y_x|c)+\delta P(y_{x'}|c)+\theta P(y'_{x'}|c) +\nonumber\\
&&(\beta -\gamma-\theta +\delta) P(y_x,y'_{x'}|c).\nonumber
\end{eqnarray}
Case 1: If $\beta -\gamma-\theta +\delta \ge 0$,
\begin{eqnarray*}
&&f(c)\\
&\le& (\gamma -\delta)P(y_x|c)+\delta P(y_{x'}|c)+\theta P(y'_{x'}|c) + \nonumber\\
&&\frac{\beta - \gamma - \theta + \delta}{2} + \frac{|\beta - \gamma - \theta + \delta|}{2}\\
&=& (\gamma -\delta)P(y_x|c)+\delta P(y_{x'}|c)+\theta P(y'_{x'}|c) +\nonumber\\
&&\beta - \gamma - \theta + \delta.
\end{eqnarray*}
and,
\begin{eqnarray*}
&&f(c)\\
&\ge& (\gamma -\delta)P(y_x|c)+\delta P(y_{x'}|c)+\theta P(y'_{x'}|c) +\nonumber\\
&& \frac{\beta - \gamma - \theta + \delta}{2} - \frac{|\beta - \gamma - \theta + \delta|}{2}\\
&=& (\gamma -\delta)P(y_x|c)+\delta P(y_{x'}|c)+\theta P(y'_{x'}|c).
\end{eqnarray*}
Therefore, $f(c)$ is $\frac{|\beta - \gamma - \theta + \delta|}{2}$-identified to $(\gamma -\delta)P(y_x|c)+\delta P(y_{x'}|c)+\theta P(y'_{x'}|c) + \frac{\beta - \gamma - \theta + \delta}{2}$.\\
Case 2: If $\beta -\gamma-\theta +\delta < 0$,
\begin{eqnarray*}
&&f(c)\\
&\le& (\gamma -\delta)P(y_x|c)+\delta P(y_{x'}|c)+\theta P(y'_{x'}|c) +\nonumber\\
&& \frac{\beta - \gamma - \theta + \delta}{2} + \frac{|\beta - \gamma - \theta + \delta|}{2}\\
&=& (\gamma -\delta)P(y_x|c)+\delta P(y_{x'}|c)+\theta P(y'_{x'}|c).
\end{eqnarray*}
and,
\begin{eqnarray*}
&&f(c)\\
&\ge& (\gamma -\delta)P(y_x|c)+\delta P(y_{x'}|c)+\theta P(y'_{x'}|c) +\nonumber\\
&& \frac{\beta - \gamma - \theta + \delta}{2} - \frac{|\beta - \gamma - \theta + \delta|}{2}\\
&=& (\gamma -\delta)P(y_x|c)+\delta P(y_{x'}|c)+\theta P(y'_{x'}|c)  +\nonumber\\
&& \beta - \gamma - \theta + \delta.
\end{eqnarray*}
Therefore, $f(c)$ is $\frac{|\beta - \gamma - \theta + \delta|}{2}$-identified to $(\gamma -\delta)P(y_x|c)+\delta P(y_{x'}|c)+\theta P(y'_{x'}|c) + \frac{\beta - \gamma - \theta + \delta}{2}$.
\end{proof}

\end{document}